\documentclass[twoside]{article}

\usepackage[turnoff]{notes}

\usepackage{wrapfig}

\usepackage{graphicx} 
\usepackage{subfigure} 
\usepackage{algorithm}
\usepackage{algorithmic}
\usepackage{color}
\usepackage{amssymb}
\usepackage{amsmath}

\newcommand{\vectt}[1]{\boldsymbol{#1}}
\newcommand{\vect}[1]{\mathbf{#1}}

\newcommand{\svm}{{\em SVM}}

\newcommand{\esvm}{{\em $\epsilon$-SVM}}
\newcommand{\lmnn}{{\em LMNN}}

\newcommand{\gammasvm}{\gamma}
\newcommand{\gammalmnn}{\gamma}
\newcommand{\gammanewsvm}{\gamma}

\usepackage[accepted]{aistats2012}
\begin{document}

%
\runningtitle{A metric learning perspective of SVM: on the LMNN-SVM relation}

%
\runningauthor{Huyen Do, Alexandros Kalousis, Jun Wang, Adam Woznica}

\twocolumn[

\aistatstitle{A metric learning perspective of SVM:\\
	on the relation of LMNN and SVM}

\aistatsauthor{Huyen Do \And Alexandros Kalousis \And Jun Wang \And Adam Woznica }

\aistatsaddress{Computer Science Dept. \\ University of Geneva\\ Switzerland\And Computer Science Dept. \\ University of Geneva\\ Switzerland\And Computer Science Dept. \\ University of Geneva\\ Switzerland \And Computer Science Dept. \\ University of Geneva\\ Switzerland} ]

\begin{abstract}
  
Support Vector Machines, \svm s, and the Large Margin Nearest Neighbor algorithm, \lmnn, are two very popular learning algorithms with quite different learning 
biases. In this paper we bring them into a unified view and show that they have a much stronger relation than what is commonly thought.
We analyze \svm s from a metric learning perspective and cast them as a metric learning problem, a view which helps us uncover the relations 
of the two algorithms. We show that \lmnn\ can be seen as learning a set of local \svm-like models in a quadratic space. Along the way and 
inspired by the metric-based interpretation of \svm s we derive a novel variant of \svm s, \esvm, to which \lmnn\ is 
even more similar. 
We give a unified view of \lmnn\ and the different \svm\ variants. Finally we provide some preliminary experiments on a number of benchmark datasets in which show that \esvm\ compares favorably both with respect to \lmnn\ and \svm.
\end{abstract}

\section{Introduction}
Support Vector Machines, \cite{Taylor2000}, and metric learning algorithms, \cite{weinberger2009distance, globerson2006mlc, goldberger2005nca,xing2003dml},
are two very popular learning paradigms with quite distinct learning biases. In this paper we focus on \svm s and \lmnn, one of the most prominent metric 
learning algorithms, \cite{weinberger2009distance}; we bring them into a unified view and show that they have a much stronger relation than what is commonly accepted.

We show that \svm\ can be formulated as a metric learning 
problem, a fact which provides new insights to it. Based on these insights and employing learning biases typically used in metric learning we 
propose \esvm, a novel \svm\ variant which is shown to empirically outperform \svm.  
\esvm\ relates, but is simpler, to the radius-margin ratio error bound optimization that is often used when \svm\ is 
coupled with feature selection and weighting, \cite{Rakotomamonjy2003,Huyen2009}, and multiple kernel learning 
\cite{Chapelle2002,Huyen2009b,GaiChenZhang2010}. 
More importantly we demonstrate a very strong and previously unknown connection between \lmnn\ and \svm. Until now \lmnn\ 
has been considered as the distance-based counterpart of \svm, in the somehow shallow sense that both use 
some concept of margin, even though their respective margin concepts are defined differently, and the hinge 
loss function, within a convex optimization problem. We show that the relation between \lmnn\ and \svm\ is 
much deeper and demonstrate that \lmnn\ can be seen as a set of local \svm-like classifiers in a quadratic 
space. In fact we show that \lmnn\ is even more similar to \esvm\ than \svm. This strong connection has the potential 
to lead to more efficient \lmnn\ implementations, especially for large scale problems and vice versa, to lead to more efficient schema of multiclass \svm. Moreover the result is also valid for other large margin metric learning algorithms.
Finally we use the metric-based 
view to provide a unified view of \svm, \esvm\ and \lmnn.

Overall the main contribution of the paper is the unified view of \lmnn, \svm\ and the variants of the latter. Along the way we also devise a new algorithm, \esvm, that combines ideas from both \svm\ and \lmnn\ and finds its support in the \svm\ error bound.

In the next section we will briefly describe the basic concepts of \svm\ and \lmnn. In section \ref{sec:SVM-metric-view} we describe 
\svm\ in the metric learning view and the new insights that this view brings. We also discuss some invariant properties of \svm\ and how metric learning may or may not help 
to improve \svm. Section \ref{ESVM} describes \esvm, a metric-learning and \svm\ inspired algorithm. Section \ref{sec:LMNN} discusses the relation of \lmnn\ and \svm, and provides a unified view of the different \svm\ and
\lmnn\ variants. 
In section~\ref{experiments} we give some experimental results for \esvm\ and compare it with \svm\ and \lmnn. Finally we conclude in section~\ref{conclusion}.

\subsection{Basic \svm\ and \lmnn\ concepts}
We consider a binary classification problem in which we are given a set of $n$ learning instances 
$S = \{(\vect{x}_{1}, y_{1}),..., (\vect{x}_{n},y_{n})\}, \vect{x}_i \in {\mathcal R}^d$, where $y_i$ is the class
label of the $\vect{x}_i$ instance and $y_i \in \{+1, -1\}$. We denote by $\|\vect{x}\|_p$ the $l_p$
norm of the $\vect{x}$ vector. Let  $H_{\vect w}$ be a hyperplane given by $\vect{w}^T\vect{x} + b=0$; the 
signed distance, $d(\vect {x}_i, H_{\vect w})$, of some point $\vect {x}_i$ from the $H_{\vect w}$ hyperplane is given by:
 $d(\vect {x}_i, H_{\vect w})=\frac{\vect{w}^T\vect{x_i} + b}{\|\vect{w}\|_p}$.
\svm s learn a hyperplane $H_{\vect w}$ which separates the two classes
and has maximum margin $\gammasvm$, which is, informally, the distance of the nearest instances from $H_{\vect w}$:
$
\gammasvm=\min_i [y_i d(\vect x_i, H_{\vect w})]_+   \label{eq:svm.margin}
$  \cite{Taylor2000}.
The optimization problem that \svm s solve is:
\begin{eqnarray}
\label{eq:SVM-std-margin}
\max_{\vect{w},b, \gamma} \,\,\,\, \gammasvm        & & 
s.t.             \,\, \,\,\,\, \frac{y_i(\vect{w}^T \vect{x}_i + b)}{\|\vect w\|_2} \geq \gammasvm,  \ \forall i\ 
\end{eqnarray}
which is usually rewritten to avoid the uniform scaling problem with $\vect w, b$, as:
\begin{eqnarray}
\label{eq:SVM-std}
\min_{\vect{w},b} \,\,\,\, \|\vect{w}\|^2_2  &&
s.t. \,\,\,\,\,\, y_i(\vect{w}^T \vect{x}_i + b) \geq 1, \forall i 
\end{eqnarray}
The margin maximization is motivated by the \svm\ error bound which is a function of the $R^2/\gammasvm^2$ ratio; $R$ is
the radius of the smallest sphere that contains the learning instances. Standard \svm s only focus on the margin and ignore
the radius because for a given feature space this is fixed. However  the $R^2/\gammasvm^2$ ratio has been used as an 
optimization criterion in several cases, for feature selection, feature weighting and multiple kernel learning \cite{Chapelle2002, Rakotomamonjy2003, Huyen2009b,Huyen2009,GaiChenZhang2010}. The biggest challenge when using the radius-margin bound is that it leads 
to non convex optimization problems. 

In metric learning we learn a Mahalanobis metric parametrized by a 
Positive Semi-Definite (PSD) matrix $\mathbf M$ under some cost function and some constraints on the Mahalanobis distances 
of same and different class instances. 
The squared Mahalanobis distance has the following form
$
d^2_{\mathbf{M}}(\vect{x}_i, \vect{x}_j)=(\vect{x}_i-\vect{x}_j)^T\vect{M}(\vect{x}_i-\vect{x}_j) 
$. 
$\mathbf M$ can be rewritten as $\mathbf M=\mathbf L^T \mathbf L$, i.e 
the Mahalanobis distance computation can be considered as a two-step procedure, that first computes a linear transformation of the instances
given by the matrix $\mathbf L$ and then the Euclidean distance in the transformed space, i.e $d^2_{\mathbf{M}}(\vect{x}_i, \vect{x}_j) = d^2(\mathbf L\vect{x}_i, \mathbf L \vect{x}_j) $.

\lmnn, one of the most popular metric learning methods, also works based on the concept of margin which is nevertheless different from that of \svm. 
While the \svm\ margin is defined \textit{globally} with respect to a hyperplane, the \lmnn\ margin is defined \textit{locally} with respect to center 
points. The \lmnn\ margin, $\gammalmnn_{\vect{x}_0}$, of a center point, instance $\vect{x}_0$, is given by:
\begin{eqnarray}
\label{eq:lmnn.margin}
\gammalmnn_{\vect{x}_0}   &=&  \min_{i,j} [d_{\mathbf M}^2(\vect{x}_0 , \vect{x}_j) -  d_{\mathbf M}^2(\vect{x}_0 , \vect{x}_i)]_+ \\
     &&y_0 \neq y_j, \vect{x}_i \in targ(\vect{x}_0)\nonumber
\end{eqnarray}
where $targ(\vect{x}_0)$ is the set of the \textit{target neighbors} of $\vect{x}_0$, which is defined as 
$targ(\vect x_0)=\{ \vect{x}_i | \vect{x}_i \in \text{ neighborhood of } {\vect x}_0 \wedge y_0==y_i\}$. The neighborhood can be 
defined either as a $k$ nearest neighbors or as a sphere of some radius. 

\lmnn\ maximizes the sum of the margins of all instances. 
The underlying idea is that it learns a metric $\mathbf M$ or a linear transformation $\mathbf L$ which brings instances close to their same class center point while it moves away from it different class instances with a margin of one. This learning bias is commonly referred as large margin metric learning \cite{Schultz2004,weinberger2009distance,Shwartz2004,Kaizhu2009}.
\lmnn\ optimization problem is \cite{weinberger2009distance}: 
\begin{eqnarray}
\label{eq:LMNN_org}
 \min_{\mathbf M, \vectt{\xi}} && \sum_{i} \sum_{\vect x_j \in targ(\vect x_i)} (d_{\mathbf M}^2(\vect x_i,\vect x_j ) + C \sum_l (1 - y_i y_l) \xi_{ijl}) \nonumber \\
s.t. &&  d_{\mathbf M}^2(\vect x_i,\vect x_l ) -  d_{\mathbf M}^2(\vect x_i,\vect x_j )  \geq 1 - \xi_{ijl},    \nonumber \\
&& \forall(j|\vect x_j \in targ(\vect x_i)), \forall i,l;  \,\,\, \vectt \xi \geq 0, \,\,\,\, \mathbf M \succeq 0  
\end{eqnarray}
\section{\svm\ under a metric learning view}
\label{sec:SVM-metric-view}
We can formulate the \svm\ learning problem as a metric learning problem in which the learned transformation matrix is diagonal. 
To do so we will first define a new metric learning problem and then we will show its equivalence to \svm. 

We start by introducing the fixed hyperplane $H_{\vect 1}$, the normal vector of which is $\vect{1}$ (a $d$-dimensional vector of ones), i.e. $H_1: \vect{1}^T\vect{x}+b=0$.
Consider the following linear transformation $\tilde {\vect{x}} = \mathbf W\vect{x}$, where $\mathbf W$ is a $d \times d$ diagonal matrix with $diag(\mathbf W)=\vect{w}=(w_1,\dots,w_d)^T$.

We now define the margin $\gammanewsvm$ of two classes with respect to the hyperplane $H_1$ as the 
minimum absolute difference of the distances of any two, different-class instances, $\tilde{\vect{x}_i}, \tilde{\vect{x}_j}$, projected to the norm vector of $H_{\vect 1}$ hyperplane,
which can be finally written as: 
$\gammanewsvm  = \min_{i,j,y_i \neq y_j} \frac{|\vect{1}^T\vect{W}( \vect{x}_i -\vect{x}_j )|}{\sqrt d}= \min_{i,j,y_i \neq y_j} |d(\tilde{\vect{x}_i}, H_{\vect 1})- d(\tilde{\vect{x}_j}, H_{\vect 1})|$. In fact this is the \svm\ margin, see equation (\ref{margin.equip}) in Appendix.

We are interested in that linear transformation, $diag(\mathbf W)=\mathbf{w}$, for which it holds $y_i(\vect{w}^T \vect{x}_i + b) \geq 0$, i.e. 
the instances of the two classes lie on two different sides of the hyperplane $H_{\vect 1}$, and which has a maximum margin.
This is achieved by the following metric-learning optimization problem \footnote{Notice that $(\vect{w}^T( \vect{x}_i -\vect{x}_j ))^2 = ( \vect{x}_i -\vect{x}_j )^T\vect{w} \vect{w}^T( \vect{x}_i -\vect{x}_j )$ is the Mahalanobis distance associated with the rank 1 matrix $\mathbf M = \vect{w}\vect{w}^T$}: 
\begin{eqnarray}
\label{SVM_metric_org}
\max_{\vect{w}} \min_{i,j,y_i \neq y_j}  && ( \vect{x}_i -\vect{x}_j )^T\vect{w}\vect{w}^T( \vect{x}_i -\vect{x}_j )  \\
 s.t.  &&  y_i(\vect{w}^T \vect{x}_i + b) \geq 0, \forall i\nonumber\\
%
\label{SVM_gamma_sqd}
\Leftrightarrow 
\max_{\vect{w}, \gamma}  &&  \gammanewsvm^2 \\
s.t.  && (\vect{w}^T( \vect{x}_i -\vect{x}_j ))^2 \geq \gammanewsvm^2, \forall (i,j,y_i \neq y_j) \nonumber\\
       && y_i(\vect{w}^T \vect{x}_i + b) \geq 0, \forall i \nonumber
\end{eqnarray}
Additionally we prefer the $\mathbf W$ transformation that places the two classes ``symmetrically'' with respect to the $H_{\vect 1}$ hyperplane, 
i.e. the separating hyperplane is at the middle of the margin instances of the two classes: $y_i(\vect{w}^T \vect{x}_i + b) \geq \gammanewsvm/2, \forall i$. 
Notice that this can always be achieved by 
adjusting the translation parameter $b$ by adding to it the $\frac{\gammanewsvm_1 - \gammanewsvm_2}{2}$ value, where $\gammanewsvm_1$ 
and $\gammanewsvm_2$ are respectively the margins of the $y_1$ and $y_2$ classes to the $H_{\vect 1}$ hyperplane. Hence the parameter 
$b$ of the $H_1$ hyperplane can be removed and replaced by an equivalent translation transformation $b$. From now on, we assume 
$H_{\vect 1}: \vect{1}^T\vect{x} + 0 = 0$ and we add a translation $b$ after the linear transformation $\mathbf W$.  With this 
'symmetry' preference, (\ref{SVM_gamma_sqd}) can be reformulated as (see the detailed proof in Appendix):
\begin{eqnarray}
\label{eq:SVMMetric_org}
\max_{\vect{w},b, \gamma} \,\, \,\, \gammanewsvm &&
s.t. \,\, \,\, y_i(\vect{w}^T \vect{x}_i + b) \geq \gammanewsvm, \forall i 
\end{eqnarray}
This optimization problem learns a diagonal transformation $\mathbf W$ and a translation $b$ which maximize the 
margin and place the classes 'symmetrically' to the $H_1: \vect{1}^T\vect{x}=0$ hyperplane. However this optimization problem scales with the  $\vect{w}$ (as is the case with the \svm\ formulation (\ref{eq:SVM-std-margin})).
\note[Removed]
{
if $(\vect{w},b)$ is a feasible solution of (\ref{eq:SVMMetric_org}) 
then $(\lambda\vect{w}, \lambda b), \forall \lambda \geq 0$, is also a feasible point with margin $\lambda \gammanewsvm$.
}
Therefore we need a way to control the uniform scaling of $\vect{w}$. 
One way is to fix some 
norm of $\vect{w}$ e.g.  $\|\vect{w}\|_p = 1$, and the optimization problem becomes:\begin{eqnarray}
\label{eq:SVMMetric_org2}
\max_{\vect{w}, \gamma,b}  \,\, \,\, \gammanewsvm &&
s.t. \,\, \,\,  y_i(\vect{w}^T \vect{x}_i + b) \geq \gammanewsvm, \forall i ,  \|\vect{w}\|_p = 1 
\end{eqnarray}
Notice that both problems~(\ref{eq:SVMMetric_org}) and~(\ref{eq:SVMMetric_org2}) are still different from the standard \svm\ formulation given in~(\ref{eq:SVM-std-margin}) or~(\ref{eq:SVM-std}). However it can be shown that they are in fact equivalent to \svm; for the detailed proof see in the Appendix.


Thus we see that  starting from a Mahalanobis metric learning problem (\ref{SVM_metric_org}), i.e learning a linear transformation $\mathbf W$, and with 
the appropriate cost function and constraints on pairwise distances, we arrive to a standard \svm\ learning problem. 
We can describe \svm\ in the metric learning jargon as follows: it learns a diagonal linear transformation $\mathbf W$ 
and a translation $b$ which maximize the margin and place the two classes symmetrically in the opposite sides 
of the hyperplane $H_{\vect1}:\vect{1}^T\vect{x}=0$. In the standard view of \svm, the space is fixed and the hyperplane is moved 
around to achieve the optimal margin. In the metric view of \svm, the hyperplane is fixed to $H_{\vect 1}$ and 
the space is scaled, $\mathbf W$, and then translated, $b$, so that the instances are placed optimally around 
$H_{\vect 1}$. Introducing a fixed hyperplane $H_{\vect 1}$ will provide various advantages in relation to the different 
radius-margin \svm\ versions and \lmnn\ as we will show soon. 
It is also worth to note that the \svm\ metric view holds true for any kernel space, since its final optimization problem is the same as that of standard \svm, i.e we can kernelize it directly as \svm.

From the metric learning perspective, one may think that learning a full linear transformation instead of a diagonal could bring more advantage, however as we will see right away this is not true for the case of \svm.
For any linear transformation $\mathbf L$ (full matrix), the distance of $\mathbf L \vect{x}$ to the 
hyperplane $H_{\vect 1}: \vect{1}\vect{x} + b = 0$ is:
$
d(\mathbf L \vect{x}, H_{\vect 1}) =\frac{\vect{1}^T \mathbf L \vect{x} + b}{\sqrt{d}} = \frac{\vect{1}^T\mathbf D_{\mathbf L} \vect{x} + b}{\sqrt{d}} 
$
where $\mathbf D_{\mathbf L} $ is a diagonal matrix, in which the $k$ diagonal element corresponds to the sum 
of the elements of the $k$th column of $\mathbf L$: ${\mathbf D_{\mathbf L}}_{kk} = \sum_i L_{ik}$. So for any full 
transformation matrix $\mathbf L$  there exists a diagonal transformation $\mathbf D_{\mathbf L}$ which has the same signed distance 
to the hyperplane. This is also true for any hyperplane $\vect{w}^T\vect{x} + b$ where $\vect{w}$ does not contain 
any zero elements. Thus learning a full matrix does not bring any additional advantage to \svm. 

\section{\esvm, an alternative to the radius-margin approach}
\label{ESVM}
\begin{wrapfigure}{r}{0.20\textwidth}
  \begin{center}
    \includegraphics[scale=0.25]{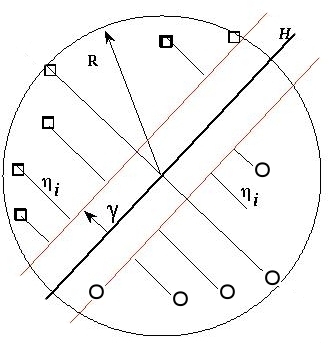}
  \caption{Within class distance: controlling the radius $R$ or the instance-hyperplane distances $\eta_i$.
 \label{fig:esvm}}
  \end{center}
\end{wrapfigure}
A common bias in metric learning is to learn a metric which keeps instances of the same class close while pushing instances of different classes far away \cite{}. This bias is often implemented through local or global constraints on the pair-wise distances of same-class and different-class instances which make the \textit{between class distance} large and the \textit{within class distance} small.
Under the metric view of \svm\ we see that the learned linear transformation does control 
the between class distance by maximizing the margin, but it ignores the within class distance. 
Inspired by the metric learning biases we will extend the \svm\ under the metric view and incorporate constraints on the within-class distance which we will minimize.

We can quantify the within class distance with a number of different ways such as the 
sum of instance distances from the centers of their class, as in Fisher Discriminant Analysis (FDA) \cite{Duda2001}, or by the pairwise distances of the instances of each 
class, as it is done in several metric learning algorithms~\cite{xing2003dml,davis2007itm,weinberger2009distance}. Yet another way
to indirectly minimize the within class distance is by minimizing the total data spread while maximizing the between class distance. 
One measure of the total data spread is the radius, $R$, of the sphere that contains the data; so by minimizing the radius-margin ratio,  
${R^2}/{\gamma^2}$, we can naturally minimize the within class distance while maximizing the between. 
Interestingly the \svm\ theoretical error bound points towards the minimization of the same quantity, i.e. the radius-margin ratio,
thus the proposed metric learning bias finds its theoretical support in the \svm\ error bound. However optimizing over the margin 
and the radius poses several difficulties since the radius is computed in a complex form, \cite{Rakotomamonjy2003,Huyen2009b,GaiChenZhang2010}.

We can avoid the problems that are caused by the radius by using a different quantity to indirectly control the within class distance. We propose to minimize instead of the radius, the sum of the instance distances from the margin hyperplane, this sum is yet another way to control the data spread. We will call the resulting algorithm, which in addition to the margin maximization also minimizes the within-class-distance through the sum of the instance distances from the margin hyperplane, \esvm. The minimization of the sum has a similar effect to the minimization of the \svm\ radius, Figure \ref{fig:esvm}. In a later section we will show that \esvm\ can be seen as a bridge between \lmnn\ and \svm.

We define the optimization problem of \esvm\ as follows: select the transformations, one linear diagonal, $diag(\mathbf W)=\vect w$, and one translation, $b$, which maximize the 
margin and keep the instances symmetrically and within a small distance from the $H_{\vect 1}$ hyperplane. This is achieved by the following optimization problem:
\begin{eqnarray}
\label{eq:esvm_org_primal}
\min_{\vect{w},b}  &&  \vect w^T \vect w  + \lambda \sum_i max(0,y_i(\vect w^T \vect x_i + b)-1) \\
&&  + C \sum_i max(0, 1- y_i(\vect w^T \vect x_i + b) )  \nonumber
\end{eqnarray}
$max(0,y_i(\vect w^T \vect x_i + b)-1)$  penalizes instances which lie on the correct side of the hyperplane but are far from the margin. $max(0, 1-y_i(\vect w^T \vect x_i + b)$ is the \svm\ hinge loss which penalize instances violating the margin. (\ref{eq:esvm_org_primal}) is equivalent to:
\begin{eqnarray}
\label{eq:esvm_org}
\min_{\vect{w},b,\vectt \xi, \vectt \eta}  && \vect w^T \vect w + \lambda \sum_i \eta_i   + C\sum_i \xi_i  \\
 s.t. &&   1 - \xi_i \leq y_i(\vect{w}^T \vect{x}_i + b) \leq 1 + \eta_i , \forall i, \vectt \xi,\vectt\eta \geq 0 \nonumber
\end{eqnarray}
where $\xi_i$ is the standard \svm\ slack variable for instance $i$ and $\eta_i$ the distance of that instance from the margin hyperplane. The dual form of 
this optimization problem is:
\begin{eqnarray}
 \max_{\vectt{\alpha},\vectt{\beta}} &\frac{-1}{2} \sum_{ij} (\alpha_i - \beta_i)(\alpha_j - \beta_j) y_iy_j \vect{x}_i \vect{x}_j + \sum_i(\alpha_i - \beta_i) \nonumber\\
 s.t. & \sum_i(\alpha_i-\beta_i) y_i = 0;  \, 0 \leq \alpha_i  \leq C ; \, 0 \leq \beta_i  \leq \lambda, \forall i 
\end{eqnarray}
Note that we have two hyper-parameters $C$ and $\lambda$, typically we will assign a higher value to the $C$ since we tolerate less 
the violations of the margin compared to a larger data spread. The two parameters control the trade off between the importance of the margin and the within data spread.
 
\section{On the relation of \lmnn\ and \svm}
\label{sec:LMNN}
In this section we demonstrate the relation between \lmnn\ and \svm.
Lets define a quadratic mapping $\vect \Phi$ that maps $\vect{x}$ to a quadratic space where $\vect \Phi(\vect{x}) = 
(x_1^2, x_2^2,\dots, x_d^2, x_i x_{j\{d\geq i > j \geq1 \}}, x_1, x_2,\dots,x_d) \in \mathcal R^{d'}$, ${d'=\frac{d(d+3)}{2}}$.
We will show that the \lmnn\ margin in the original space is the \svm\ margin in this quadratic space.

Concretely, the squared distance of an instance $\vect{x}$ from a center point instance $\vect x_l$
in the linearly transformed space that corresponds to the Mahalanobis distance $\mathbf M$ learned by \lmnn\ 
can be expressed in a linear form of $\vect \Phi (\vect x)$. Let $\mathbf L$ be the linear transformation 
associated with the learned Mahalanobis distance $\mathbf M$. We have:
\begin{eqnarray}
\label{lin.equivalent}
d^2_{\mathbf M}(\vect x, \vect x_l) & = & d^2(\mathbf L \vect x, \mathbf L \vect x_l) \\
&  = &\vect x^T \mathbf L^T \mathbf L \vect{x} - 2 \vect{x}_l^T \mathbf {L}^T \mathbf L \vect{x} + \vect{x}_l^T\mathbf L^T \mathbf L \vect{x}_l \nonumber\\
& = & \vect{w}_{l}^T \vect \Phi (\vect x) + b_{l} \nonumber
\end{eqnarray}
where $b_{l} = \vect{x}_l^T \mathbf L ^T \mathbf L  \vect{x}_l$ and $\vect{w}_{l}$ has two parts, quadratic $\vect w^{quad}$ and linear $\vect w_l^{lin}$:  $\vect{w}_{l}= (\vect w^{quad}, \vect w_l^{lin})$ 
where $\vect w^{quad}$ is the vector of 
the coefficients of the quadratic components of $\vect \Phi(\vect x)$,
with $d'-d$ elements given by the elements of $\mathbf L^T\mathbf L$, and $\vect w_l^{lin}$ is equal to
$-2\mathbf L^T\mathbf L \vect{x}_l$--- the vector of coefficients of the linear components of $\vect \Phi(\vect x)$.
$d^2(\mathbf L \vect x, \mathbf L \vect x_l)$ is proportional to the distance of $\vect \Phi({\vect{x}})$ from the 
hyperplane $H_l: \vect{w}_{l}^T \vect \Phi(\vect{x}) + b_l  = 0$ since $d^2(\mathbf L \vect x, \mathbf L \vect x_l)=\|\vect w_{l}\| d(\vect{\Phi}(\vect x),H_l)$.  
Notice that this value is always non negative, so in the quadratic space $\vect \Phi (\vect{x})$ always lies on the positive side of $H_l$.

A more general formulation of the \lmnn\ optimization problem~(\ref{eq:LMNN_org}) which reflects the same learning bias of \lmnn, i.e for each center point, keeps the same-class instances close to the center and pushes different-class instances outside the margin, is:
\begin{eqnarray}
\label{eq:LMNN_general}
 \min_{\mathbf M, \vectt{\xi}, \gamma_{\vect x_l, \forall l} } && \sum_{l} \frac{1}{ \gamma^2_{\vect x_l}} +  \lambda \sum_{l} \sum_{\vect x_i \in targ(\vect x_l)} (d_{\mathbf M}^2(\vect x_i,\vect x_l ) \nonumber\\
&  + & C  \sum_j (1 - y_j y_l) \xi_{ijl})  \\
s.t. &&   d_{\mathbf M}^2(\vect x_j,\vect x_l ) -  d_{\mathbf M}^2(\vect x_i,\vect x_l )  \geq \gamma_{\vect x_l} - \xi_{ijl}, \nonumber \\
&&  \forall(i|\vect x_i \in targ(\vect x_l)), \forall j,l; \nonumber\\
&& \gamma_{\vect x_l} \geq 0, \forall l, \,\,\,\, \vect \xi \geq 0, \,\,\,\, \mathbf M \succeq 0 \nonumber
\end{eqnarray}

\begin{figure}[ht]
\begin{minipage}[b]{1\linewidth}
\centering
\subfigure[Standard \lmnn\ model view]{
\includegraphics[scale=0.25]{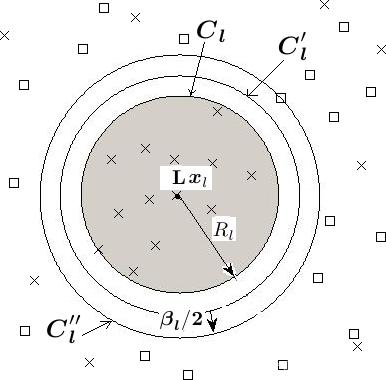}
\label{fig:LMNN}
}
\subfigure[\lmnn\ model view under an \svm-like interpretation]{
\includegraphics[scale=0.25]{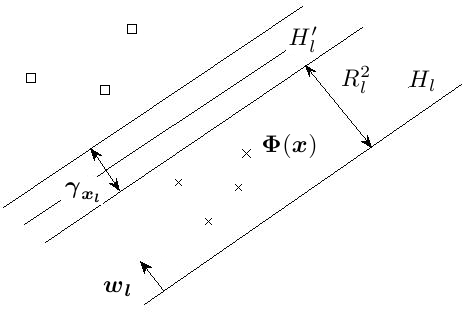}
\label{fig:LMNN_SVM}
}
\caption{Alternative views on \lmnn }
\label{fig:views1}
\end{minipage}
\end{figure}

Standard \lmnn\ puts one extra constraint on each $\gamma_{\vect x_l}$, it sets each of them to one. With this constraint problem~(\ref{eq:LMNN_general}) reduces to (\ref{eq:LMNN_org}).
We can rewrite problem~(\ref{eq:LMNN_general}) as: 
\begin{small}
\begin{eqnarray}
\label{LMNN_general2}
\min_{\mathbf L,\vectt{\xi},  R_{l, \forall l}, \gamma_{\vect x_l ,\forall l}} &&  \sum_l  \frac{1}{\gammalmnn^2_{\vect x_l}}  +  C \sum_{li} \xi_{li}   \\
&+&   \lambda \sum_{l}\sum_{\vect x_i \in targ(\vect x_l)} d^2(\mathbf L \vect{x}_i, \mathbf L \vect{x}_l) \nonumber \\
s.t.\,\, d^2(\mathbf L\vect{x}_i, \mathbf L\vect{x}_l)  &\leq&    R_l^2 + \xi_{li}, \forall \vect{x}_i \in targ(\vect{x}_l); \forall \vect{x}_l   \nonumber\\ 
d^2(\mathbf L\vect{x}_j, \mathbf L\vect{x}_l)  &\geq &   R_l^2 + \gammalmnn_{\vect x_l} - \xi_{lj}, \forall( \vect{x}_j| y_j \neq y_l); 
\forall \vect{x}_l, \nonumber\\ 
&& \gamma_{\vect x_l} \geq 0, \forall l, \vectt \xi \geq 0 \nonumber
\end{eqnarray}  
\end{small}
In this formulation, visualized in fig.~\ref{fig:views1}\subref{fig:LMNN}, the target neighbors (marked with $\times$) of the $\vect x_l$ 
center point are constrained within the $C_l$ circle with radius $R_l$ and center $\mathbf L \vect x_l$
while the instances that have a label which is different from that of $\vect x_l$ (marked with $\Box$) 
are placed outside the circle $C_l''$ with center also  $\mathbf L \vect x_l$
and radius $R_l'' = \sqrt{R_l^2 + \gamma_{\vect x_l}}$. We denote by $\beta_l$ the difference $R''_l - R_l$ of the radii of the two circles.

\begin{figure}[ht]
\begin{minipage}[b]{1\linewidth}
\centering
\subfigure[\lmnn\ in a local \svm-like view]{
\includegraphics[scale=0.23]{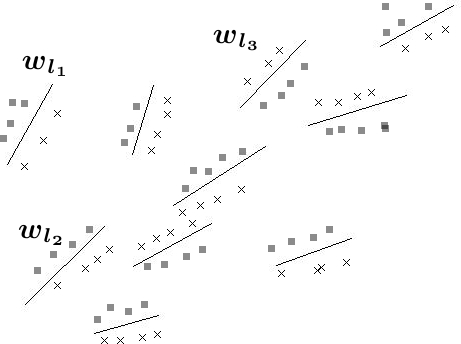}
\label{fig:localSVM}
}
\subfigure[\lmnn\  in an \svm\  metric learning view]{
\includegraphics[scale=0.23]{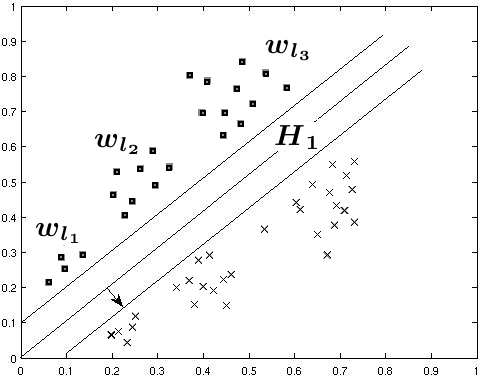}
\label{fig:SVMmetric}
}
\caption{On the relation of \lmnn\ and \svm}
\label{fig:views2}
\end{minipage}
\end{figure}

If we replace all the $d^2(\mathbf L\vect{x}, \mathbf L\vect{z})$ terms in problem~(\ref{LMNN_general2}) with their linear equivalent 
in the quadratic space, problem~(\ref{lin.equivalent}), and break up the optimization problem to a series of $n$ optimization 
problems one for each instance $\vect x_l$ then we get for each $\vect x_l$ the following optimization problem:
\begin{small}
\begin{eqnarray}
\label{eq:svm.view.of.lmnn}
 \min_{\vect w_{l}, b_l',  \vectt \xi_l, \gammalmnn_{\vect x_l}}   &&    \frac{1}{\gammalmnn^2 _{\vect x_l}} + C \sum_{\vect x_i \in B(\vect x_l)} \xi_{li}\\
&+& \lambda \sum_{\vect{x}_i \in  targ(\vect{x}_l)} (\vect{w}_{l}^T(\vect \Phi(\vect{x}_i) -  \vect \Phi(\vect{x}_l)))  \nonumber\\
 s.t. \,\,\vect{w}_{l}^T\vect \Phi(\vect{x}_i) + b'_{l}  &\leq&  -\gammalmnn_{\vect x_l}/2 +  \xi_{li}, \forall \vect{x}_i \in targ(\vect{x}_l) \nonumber\\
       \vect{w}_{l}^T\vect \Phi(\vect{x}_j) + b'_{l} & \geq & \gammalmnn_{\vect x_l}/2 - \xi_{lj},  \forall \vect{x}_j,  y_j \neq y_l \nonumber \\
     \,\,  \vect{w}_{l}^T(\vect \Phi(\vect{x}_i) -\vect \Phi(\vect{x}_l)) &\geq&  0,\forall \vect x_i \in B(\vect x_l), \,\,\,  \gamma_{\vect x_l} \geq 0, \vectt \xi_l \geq 0 \nonumber
\end{eqnarray} 
\end{small}
\noindent where $\vect w_l$ is not an independent variable but its quadratic and linear components are related as described above (formula (\ref{lin.equivalent})). The instance set $B(\vect x_l)=\{ \vect x_i \in targ(\vect x_l)\} \cup \{ \vect x_i | y_i \neq y_l\} $ is the set of all target neighbors and different class instances. In the formula (\ref{eq:svm.view.of.lmnn}) we replace $b_l$ by $-\vect{w}_{l}^T\vect \Phi(\vect{x}_l)$ since $\vect{w}_{l}^T\vect \Phi(\vect{x}_l)+b_l = 0$, and $b'_{l} = b_{l} -(R_l^2 + \gammalmnn_{\vect x_l}/2)$. We denote the hyperplane $\vect{w}_{l}^T\vect \Phi(\vect{x}) + b'_{l}=0$  by $H'_l$, as in fig.~\ref{fig:views2}. 

Now let $y_{li}:= y_l y_i$, so  $y_{li} = 1, \forall \vect x_i \in targ(\vect x_l)$ and $y_{lj} = -1, \forall (\vect x_j|y_i \neq y_l) $.  Therefore 
$- y_{li}(\vect{w}_{l}^T\vect \Phi(\vect{x}_i) + b'_{l})  \geq \gammalmnn_{\vect x_l}/2 -  \xi_{li}, \forall \vect x_i \in B(\vect x_l)$.
Without loss of generality we can assume that $y_l = -1$ and problem~(\ref{eq:svm.view.of.lmnn}) becomes: 
\begin{small}
\begin{eqnarray}
\label{eq:svm.view.of.lmnn2}
 \min_{\vect w_{l},b_l', \vectt \xi_l, \gammalmnn_{\vect x_l}}   &&   \frac{1}{\gammalmnn^2_{\vect x_l}}  + C \sum_{\vect x_i \in B(\vect x_l)} \xi_{li} \\
&+ &\lambda \sum_{\vect{x}_i \in targ(\vect{x}_l)} (\vect{w}_{l}^T(\vect \Phi(\vect{x}_i) -\vect \Phi(\vect{x}_l)) \nonumber\\
 s.t. \,\,\, y_{i}(\vect{w}_{l}^T\vect \Phi(\vect{x}_i) + b'_{l})  &\geq& \gammalmnn_{\vect x_l}/2 -  \xi_{li}, \forall \vect x_i \in B(\vect x_l) \nonumber \\
  \vect{w}_{l}^T(\vect \Phi(\vect{x}_i) -\vect \Phi(\vect{x}_l)) &\geq& 0,\forall \vect x_i \in B(\vect x_l), \,\,\, \gamma_{\vect x_l} \geq 0, \vectt \xi_l \geq 0 \nonumber
\end{eqnarray} 
\end{small}
where $\vect w_l$ is still constrained as in~(\ref{eq:svm.view.of.lmnn}). It is worth to note that the second constraint of (\ref{eq:svm.view.of.lmnn2})
will ensure that $d^2_{\mathbf M}(\vect{x}_i, \vect x_l)$ is bigger than or equal to 0, i.e it will \textit{almost} ensure the PSD of the matrix $\mathbf M$ \footnote{
\textit{Almost} in the sense that the constraint  ensures that $(\vect x-\vect x_l)^T \mathbf M (\vect x-\vect x_l) \geq 0$ for all $\vect x, \vect x_l$ in the training dataset, but can not guarantee the same for a new instance $\vect x$; to ensure $\mathbf M$ is PSD, we need $\vect z^T \mathbf M \vect z \geq 0$  for \textit{all} $\vect z$.}. However due to the specific structure of $\vect w_l$, i.e. quadratic and linear part, the PSD constraint of $\mathbf M$ is guaranteed for any $\vect x$. 

We can think of problem~(\ref{eq:svm.view.of.lmnn2}) as an extension of an \svm\ optimization problem. 
Its training set is the $B(\vect x_l)$ set, i.e. the target neighbors of $\vect x_l$ and all instances with different class label from $\vect x_l$.  
Its cost function includes, in addition to the term that 
maximizes the margin, also a sum term which forces the target neighbors of $\vect x_l$ to have small $\vect{w}_{l}^T\vect \Phi(\vect{x}_i) + b_{l}$ 
values, i.e. be at small distance from the $H_l$ hyperplane.

Minimizing the target neighbor distances from the $H_l$ hyperplane makes the distance between support vectors and $H_l$ small. It therefore 
has the effect of bringing the negative margin hyperplane of $H_l'$ close to $H_l$, bringing thus also the target neighbors close to the negative 
margin hyperplane.  In other words the optimization problem favors a small width for the band that is defined by 
$H_l$ and the negative margin hyperplane described above which contains the target neighbors.

There is an even closer relation of \lmnn\ with a local \esvm\ applied in the quadratic space that we will describe by 
comparing the optimization problems given in (\ref{eq:esvm_org_primal}) and (\ref{eq:svm.view.of.lmnn2}). \esvm\ has 
almost the same learning bias as \lmnn, the former maximizes the margin and brings all instances close to the margin hyperplanes, 
the latter also maximizes the margin but only brings the 
target neighbors close to the margin hyperplane. For each center point $ \vect x_l$ the \lmnn\ formulation, problem (\ref{eq:svm.view.of.lmnn2}), 
is very similar to that of \esvm, problem~(\ref{eq:esvm_org_primal}), applied in the quadratic space. 
The second term of the cost function of \esvm\ together with its constraints force all instances to be close to their respective margin hyperplane, this can be seen
more clear in the minimization of the $\sum \eta_i$ in problem~(\ref{eq:esvm_org}). In 
\lmnn, problem (\ref{eq:svm.view.of.lmnn2}), the third term of the cost function plays a similar role by forcing only the target neighbors to be close to the
$H_l$ hyperplane. This in turn has the effect, as mentioned above, to reduce the width of the band containing them and bringing them closer to their margin hyperplane; 
unlike \esvm\ no such constraint is imposed on the remaining instances.
%
%
So \lmnn\ is equivalent to $n$ local, modified, \esvm s, one for each instance. 
\esvm\ controls the within class distance using the distance to the margin hyperplane and  \lmnn\ using the distance to the $H_l$ hyperplane. 

Note that the $n$ optimization problems are not independent. The different 
solutions $\vect{w}_{l}$ have a common component which corresponds to the  $d(d+1)/2$ quadratic features and is given by
the $\vect w^{quad}$ vector. The linear component of $\vect w_l$, $\vect w_l^{lin}$, is a function of the specific $\vect x_l$, 
as described above. Overall learning a transformation $\mathbf L$ with \lmnn, eq.~(\ref{eq:LMNN_org}), is equivalent to learning a local \svm-like model, given by $H'_l$,
for {\em each} $\vect x_l$ center point in the quadratic space according to problems~(\ref{eq:svm.view.of.lmnn},\ref{eq:svm.view.of.lmnn2}).
%
Remember that the $\vect{w}_l, b_l$ of the different center points are related, i.e. their common quadratic part is $\vect w^{quad}$. 
\note[Removed]{
Back to standard \lmnn, where $\gamma_{\vect x_l}$ is fixed to 1 for all $\vect x_l$, replacing $b_l$ by $-\vect w^T\vect \Phi(\vect{x}_l)$, we can derive the optimization problem of \lmnn\ based on problem~(\ref{eq:svm.view.of.lmnn2}) as follows:
\begin{small}
\begin{eqnarray}
\label{eq:LMNN_esvm_full}
 \min_{\vect w_{l},b_l', \vect \xi_l}     \sum_l \sum_{\vect{x}_i \in targ(\vect{x}_l)} (\vectt \alpha^T(\vect \Phi^q (\vect{x}_i)- \vect \Phi^q (\vect{x}_l))  &&\\
+ (\vect{w}_l^{(lin)})^T(\vect \Phi^{lin}(\vect{x}_i)- \vect \Phi^{lin}(\vect{x}_l))) + C \sum_{\vect x_i \in B(\vect x_l)} \xi_{li} && \nonumber\\
 s.t. \,\,\,\, - y_{il}(\vect{w}_{l}^T\vect \Phi(\vect{x}_i) + b'_{l})  \geq 1/2 -  \xi_{li}, \forall \vect x_i \in B(\vect x_l), \forall l &&\nonumber \\
    \,\,\,  (\vect{w}_{l}^T\vect \Phi(\vect{x}_i) - \vect{w}_{l}^T\vect \Phi(\vect{x}_l) )  \geq 0, \forall \vect x_i \in targ(\vect x_l), \,\,\, \vect \xi_l \geq 0 , \forall l &&\nonumber
\end{eqnarray}
\end{small}
%
If in problem~(\ref{eq:LMNN_esvm_full})
}
If  we constrain $\vect{w}_l, b_l$ to be the same for all $\vect{x}_l$ and drop the  PSD constraints on $\vect{w}_l$ then we get the (global) \esvm-like solution. 
\note[Removed]{
Problem~(\ref{eq:LMNN_esvm_full}) can lead to more efficient \lmnn\ implementations, especially for large scale problems,
by exploiting available large scale \svm\ solvers. 
}

\paragraph{Visualization:} In fig.~\ref{fig:views1}\subref{fig:LMNN_SVM} we give a visualization of problem~(\ref{eq:svm.view.of.lmnn}) in the quadratic space $\vect \Phi(\vect{x})$. 
Figure \ref{fig:views1}\subref{fig:LMNN_SVM} gives the equivalent linear perspective in the quadratic space of the \lmnn\ model in the original space given in fig.~\ref{fig:views1}\subref{fig:LMNN}.
The center $\mathbf L \vect{x}_l$ of the $C_l$ circle in fig.~\ref{fig:views1}\subref{fig:LMNN} corresponds to 
the $H_l$ hyperplane in fig.~\ref{fig:views1}\subref{fig:LMNN_SVM}; the $C_l'$ circle with center $\mathbf L \vect{x}_l$ and radius 
$R_l' = R_l + \beta_l/2$ corresponds to the $H'_l$ separating hyperplane in fig.~\ref{fig:views1}\subref{fig:LMNN_SVM}.
Figure~\ref{fig:views2}\subref{fig:localSVM} illustrates the different local linear models in the quadratic space.
We can combine these different local models by employing the metric learning view of \svm\ 
and make the relation of \lmnn\ and \svm\ even more clear. Instead of having many local \svm-like 
hyperplanes we bring each point $\vect \Phi(\vect x_l)$ around the $H_{\vect 1}:\vect{1} \vect{x}+0 = 0$ 
hyperplane by applying to it first a $\mathbf W_l$ diagonal transformation,  $\mathbf W_l=\vect w_l$, and then a translation (fig.~\ref{fig:views2}\subref{fig:SVMmetric}). 
As before with $\vect w_l$ the different $\mathbf W_l$ transformations have a common component, which corresponds to the first $d(d+1)/2$ elements 
of the diagonal associated with the quadratic features, given by the $\vect w^{quad}$ vector, and an instance dependent component that corresponds 
to the linear features which is given by  $\vect w_l^{lin} = -2\mathbf L^T\mathbf L\vect{x}_l$; thus the translation transformation is also a function of the specific point $\vect x_l$. 
Notice that the common component has many more elements than the instance specific component. 
There is an analogy to multitask learning where models are learned over different tasks---datasets are forced to have parts of their models the same.


\paragraph{Prediction phase}
Typically after learning a metric through \lmnn\ a k-NN algorithm is used to predict the class of a new instance. Under the interpretation of \lmnn\ as a set of local \esvm\ linear classifiers this is equivalent to choosing the $k$ local hyperplanes $H_l':\vect w^T_l \vect \Phi(\vect x)+b_l'=0$ which are farther away from the new instance and which leave both the new instance and their respective center points on the same side. The farther away the new instance is from an $H_l'$ local hyperplane the closer it is to the center point associated to $H_l'$. In effect this means that we simply choose those hyperplanes---classifiers which are more confident on the label prediction of the new instance, and then we have them vote. In a sense this is similar to an ensemble learning schema 
in which each time we want to classify a new instance, we select those classifiers that are most confident. Then we have them vote in order to determine the final prediction.

\section{A unified view of \lmnn, \svm\ and its variants}
\label{sec:framework}
\begin{figure}
  \begin{center}
    \includegraphics[scale=0.47]{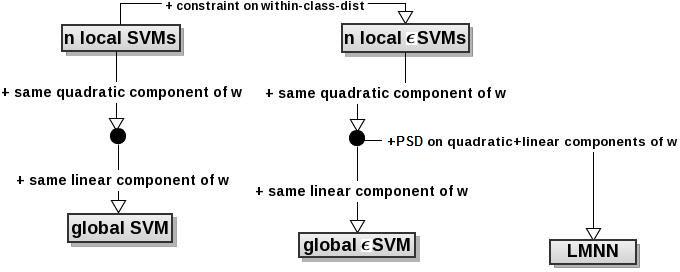}
  \caption{\begin{small}Relation among local \svm, local \esvm, global \svm, global \esvm\ and \lmnn\ in the quadratic space. \end{small}
 \label{fig:local}}
  \end{center}
\end{figure}
The standard \svm\ learning problem only optimizes the margin. On the other hand, \lmnn, as well as the different variants of radius-margin based \svm\ under the metric learning 
view (briefly mentioned in section~\ref{ESVM}), and \esvm, have additional regularizers that control directly or indirectly the within class distance. 
\lmnn\ is much closer to a collection of local \esvm s than to local \svm s, since both \lmnn\ and \esvm\ have that additional regularizer of the within 
class distance. We briefly summarize the different methods that we discussed:

\begin{list}{\labelitemi}{\leftmargin=1em}
\setlength{\itemsep}{0cm}%
\setlength{\parskip}{0cm}%
\item In standard \svm\ the space is fixed and the separating hyperplane is moved to find an optimal position in which the margin is maximized.
\item In standard \lmnn\ we find the linear transformation that maximizes the sum of the local margins of all instances while keeping the neighboring 
instances of the same class close.
\item In the metric learning view of \svm\ we find a linear diagonal transformation $\mathbf W, diag(\mathbf W) = \vect w$ plus a translation $b$ 
that maximize the margin to the fixed hyperplane $H_{\vect 1}$, clamping the norm of $\vect{w}$ to 1, or maximize the norm of $\vect{w}$ while clamping 
the margin to 1. 
\item In \esvm\ we find a linear diagonal transformation and a translation that maximize the margin with respect to the fixed hyperplane $H_{\vect 1}$ 
and keep all instances close to $H_{\vect 1}$. 
\item Finally in the interpretation of \lmnn\ as a collection of local \svm-like classifiers we find for {\em each} instance $\vect x$
a linear transformation and a translation that clamp the margin to one and keep the target neighbors of $\vect x$ within a band 
the width of which is minimized. These transformations have common components as described in the previous section. \lmnn\ is 
set of local related \esvm s in the quadratic space.
\end{list}
We will now give a high level picture of the relations of the different methods in the quadratic space $\vect \Phi(\vect x)$ by showing how from one method we can move to another by adding or dropping constraints; the complete set of relations is given in Figure \ref{fig:local}. 

We start with $n$ local non-related \svm s, if we add a constraint on the within-class distance to each of them we get $n$ non-related \esvm s. If we add additional constrains that relate the different local \esvm s, namely by constraining their quadratic components $\vect w^{quad}$ to be the same and PSD, and their linear component $\vect w_l^{lin}$ to be a function of $\vect w^{quad}$ and the center point $\vect x_l$ then we get \lmnn. 

If we go back to the original set of non-related \svm s and add a constraint that forces all of them to have the same $\vect w_l$ and $b_l'$ we get the standard global \svm. Similarly if in the collection of the local \esvm s we add a constraint that forces all of them to have the same optimal hyperplane then we also get the global \esvm. In that case all the component classifiers of the ensemble reduce to one classifier and no voting is necessary. 

\lmnn\ constrains the $n$ local \esvm s to have the same quadratic component, $\vect w^{quad}$, makes the linear components dependent on $\vect w^{quad}$ and their respective center points, accommodating like that the local information, and constrains the quadratic components $\vect w^{quad}$ to reflect the $\mathbf L^T \mathbf L$ PSD matrix, equation (\ref{lin.equivalent}). On the other hand both the global \svm\ and \esvm\ also constrain their quadratic components to be the same but remove the constraint on the PSD property of $\vect w^{quad}$; both constrain all their linear components to be the same, thus they do not incorporate local information. The last constraint is much more restrictive than the additional constrains of \lmnn, as a result the global \svm\ and \esvm\ models are much more constrained than that of \lmnn. Although we demonstrate the relations only for the case of \svm\ and \lmnn\ they also hold for all other large margin based metric learning algorithms \cite{Schultz2004,weinberger2009distance,Shwartz2004,Kaizhu2009}. 

\esvm\ builds upon two popular learning biases, namely large margin learning and metric learning and exploits the strengths of both. In addition it finds theoretical support in the radius-margin SVM error bound. \esvm\ can be seen as a bridge which connects \svm\ and existing large margin based metric learning algorithms, such as \lmnn.

\section{Experiments and results}
\label{experiments}

\begin{table*}[t]
\caption{\begin{small}Classification error. D gives the number of features of each dataset and N the number of examples.
A {\bf bold entry} in \esvm\ or \svm\ indicates that the respective method had a lower error, similarly for an {\em italics entry} for the \esvm\ and \lmnn\ pair. \end{small}\label{class_error}}
\begin{center}
\scalebox{0.68}{
\begin{tabular}{l|l|ccc|l|ccc}
\hline
  Kernel &Dataset	&\svm    & \esvm   	& \lmnn\  &  Dataset &\svm    & \esvm   	& \lmnn\ \\
\hline
	  
linear& N=62 		&   17.74  &  {\em 17.74}       & \note[rem]{NA-} 20.97  & N=198     &  25.25          &  {\bf \em 21.21} & \note[rem]{29.29-} 31.31 	\\ 
poly2 & D=2000 		&   35.48  &  35.48  	        & 35.48 	         & D=34      & 	24.24          &  {\bf \em 23.74} & 29.80 	\\ 
poly3 & &   35.48       &  {\bf \em 24.19}   & 27.42	                         &           &  {\bf 20.20}    &  {\em 20.71}     & 31.31 	\\ 
gauss1&  colonTumor     &   35.48  &  35.48  	        & 35.48                  &  wpbc &  23.74          &  23.74	          & 23.74\\ 
	\hline  
	  
linear& N=60 		&   40.00  &  {\bf \em 35.00}  &\note[rem]{NA-}45.00  	& N=351   &  {\bf 7.98}     & 11.97            & \note[rem]{\textbf{10.83}-} {\em 8.26}		\\ 
poly2 & D=7129 		&   35.00  &  35.00            &35.00                    &D=34        &   5.98          & {\em 5.98}       & 7.12		\\ 
poly3 &                 &   36.67  &  36.67            &{\em 33.33}              &            &   6.27          & 6.27             & {\em   5.98} 		\\ 
gauss1&  centralNS  	&   35.00  &  35.00            &35.00                    &  ionosphere&  10.82          & {\bf \em 5.41 }  & 15.38		\\ 
	\hline  
	  
linear & N=134 	        &   10.45  &  {\bf \em 6.72}  & \note[rem]{NA-}11.94    & N=569      &   {\bf    2.28}	&   {\em 3.69}    &\note[rem]{3.87-}4.04\\ 
poly2  & D=1524         &   60.45  &  {\bf \em 53.73} &         54.48           & D=30       &   {\bf    3.51}  &   4.75 	  &{\em 4.22}	\\ 
poly3  &                &   29.85  & {\bf \em 22.39}  & 29.85                   &            &   2.64  	        & {\bf \em 2.28 } &3.34 	\\ 
gauss1 & femaleVsMale   &   60.45  &  60.45  	      & 60.45                   &  wdbc	     &   16.34   	& {\bf \em 6.85}  &11.25	\\ 
	\hline  
linear & N=72 	        &  1.39  & 1.39	              &\note[rem]{NA-}1.39  	& N=354   &  {\bf 31.01}  &  {\em 31.30}  	  & \note[rem]{33.91-} 36.23  		\\ 
poly2  & D=7129	        &  34.72 & {\bf \em 31.94}    &        33.33   	        & D=6    &  30.72        & {\bf \em 29.86} & 36.23  	\\ 
poly3  &                &  34.72 & {\bf \em 12.50}    &        33.33 	        & &  {\bf 29.57}  & {\em 30.43}     & 31.59	\\ 
gauss1 & Leukemia       &  34.72 & 34.72              &34.72                    & 	liver	          &  {\bf 32.17} 	& {\em 32.46}     & 38.26		\\ 
	\hline  
	  
linear & N=208 	&   32.69   &  {\bf    23.08}  & \note[rem]{\em 10.10-}{\em 12.02}& N=476    &   17.02   	&  {\bf 15.76}   & \note[rem]{\textbf{4.62}-}{\em 4.83 } 	\\ 
poly2  & D=60 	&   19.23   &  {\bf    17.31}  & {\em 13.94}                    & D=166      &   {\bf 6.30}     &  6.93 	 & {\em 3.99 }	\\ 
poly3  &        &   13.46   &  {\bf    12.98}  & {\em 12.50}                    & &   {\bf 4.20}     &  {\em 4.83}    &  5.25		\\ 
gauss1 & sonar 	&   42.79   &  {\bf\em 34.62}  & 42.79                          &	musk1           &    43.49   	&  {\bf \em 39.92}&   43.07	\\ 
	\hline  

\end{tabular}
}
\end{center}
\end{table*}
In addition to studying in detail the relation between the different variants of \svm\ and metric learning we also examine
the performance of the \esvm\ algorithm. 
We experiment with \esvm, eq.~(\ref{eq:esvm_org}), and compare its performance to that of a
standard $l_1$ soft margin \svm, and \lmnn, with the following kernels:  linear, polynomial 
with degree 2 and 3, Gaussian with $\sigma = 1$. 
$C$ is chosen with 10-fold inner Cross-Validation from the set \{0.1, 1, 10, 100, 1000\}. For \esvm\ we choose to set $\lambda$ to  $C/3$ reflecting the
fact that we tolerate less the margin violations than a larger distance form the margin. We used ten 
datasets mainly from the UCI repository \cite{uci}. 
Attributes are standardized to have zero mean and one variance; kernels are normalized to have a trace of one. The number of  
target neighbors for \lmnn\ is set to 5 and its $\gamma$ parameter is set to 0.5, following the default settings suggested in \cite{weinberger2009distance}. We estimated the classification error using 10-fold CV.
The results are given in Table~\ref{class_error}.  
\begin{table}[t]
\begin{center}
\caption{\begin{small}McNemar score. loose: 0, win: 1, equal 0.5 \end{small} \label{score}}
\scalebox{0.66}{
\begin{tabular}{l|ccc}
\hline

  Kernel	&\svm    & \esvm   	& \lmnn\ \\
\hline
linear 		& 9 & 10 & 11 \\
poly2 		& 10 & 10 & 10 \\
poly3 		&10 & 11.5 & 8.5\\
gauss1 		&8.5 & 14.5 & 7\\	  
\hline

Total 		& 37.5 & 46 & 36.5  \\
\hline
\end{tabular}
}
\end{center}
\end{table}
Overall each algorithm is applied 40 times (four kernels $\times$ ten datasets). Comparing \esvm\ with \svm\ we see that the former has a lower
error than \svm\ in 19 out of the 40 applications and a higher error in only nine. A similar picture appears in the \esvm, \lmnn, pair 
where the former has a lower error in 20 out of the 40 applications and a higher in only eight. If we break down 
the comparison per kernel type, we also see that, i.e. \esvm\ has a systematic advantage over the other two algorithms no matter which kernel we use. 

%
%


To examine the statistical significance of the above results we use the McNemar's test of significance, with a significance level of 0.05. Comparing algorithms A and B on a fixed dataset and a fixed kernel algorithm $A$ is credited with one point if it was significantly better than algorithm $B$, 0.5 points if there was no significance difference between the two algorithms, and zero points otherwise. In Table \ref{score} we report the overall points that each algorithms got and in addition we break them down over the different kernels. \esvm\ has the highest score with 46 points followed by \svm\ with 37.5 and \lmnn\ with 36.5. 
The advantage of \esvm\ is much more pronounced in the case of the Gaussian kernel. This could be attributed to its additional regularization on the within-class distance which makes it more appropriate for very high dimensional spaces.

\section{Conclusion}
\label{conclusion}
In this paper, we have shown how \svm\ learning can be reformulated as a metric learning problem. 
Inspired by this reformulation and the metric learning biases, we proposed \esvm, a new \svm-based algorithm in which, in addition to the 
standard \svm\ constraints we also minimize a measure of the within class distance. More importantly the metric learning view of \svm\ helped 
us uncover a so far unknown connection between the two seemingly very different learning paradigms of \svm\ and \lmnn.  \lmnn\ can 
be seen as a set of local \svm-like classifiers in a quadratic space, 
and more precisely, as a set of local \esvm-like classifiers. Finally preliminary results show a superior performance of \esvm\ compared to both \svm\ and \lmnn. 
Although our discussion was limited to binary classification, it can be extended to 
the multiclass case. Building on the \svm-\lmnn\ relation, our current work focuses on the full analysis of the multiclass case, 
a new schema for multiclass \svm\ which exploits the advantages of both \lmnn\ and kNN in multiclass problems, and finally the exploration of 
the learning models which are in between the \svm, \lmnn\ models.



\bibliography{referenceICML2011}

\begin{thebibliography}{10}
\providecommand{\url}[1]{\texttt{#1}}
\providecommand{\urlprefix}{URL }

\bibitem{Chapelle2002}
Chapelle, O., Vapnik, V., Bousquet, O., Mukherjee, S.: Choosing multiple
  parameters for support vector machines. In: Machine Learning. vol.~46, pp.
  131--159. Kluwer Academic Publishers, Hingham, MA, USA (2002)

\bibitem{Taylor2000}
Cristianini, N., Shawe-Taylor, J.: An introduction to Support Vector Machines.
  Cambridge University Press (2000)

\bibitem{davis2007itm}
Davis, J., Kulis, B., Jain, P., Sra, S., Dhillon, I.: {Information-theoretic
  metric learning}. In: Proceedings of the 24th international conference on
  Machine learning. ACM New York, NY, USA (2007)

\bibitem{Huyen2009b}
Do, H., Kalousis, A., Hilario, M.: Feature weighting using margin and radius
  based error bound optimization in svms. In: ECML (2009)

\bibitem{Huyen2009}
Do, H., Kalousis, A., Wozica, A., Hilario, M.: Margin radius based multiple
  kernel learning. In: ECML (2009)

\bibitem{uci}
Frank, A., Asuncion, A.: {UCI} machine learning repository (2010),
  \url{http://archive.ics.uci.edu/ml}

\bibitem{GaiChenZhang2010}
Gai, K., Chen, G., Zhang, C.: Learning kernels with radiuses of minimum
  enclosing balls. In: NIPS (2010)

\bibitem{globerson2006mlc}
Globerson, A., Roweis, S.: {Metric learning by collapsing classes}. In:
  Advances in Neural Information Processing Systems. vol.~18. MIT Press (2006)

\bibitem{goldberger2005nca}
Goldberger, J., Roweis, S., Hinton, G., Salakhutdinov, R.: {Neighbourhood
  components analysis}. In: Advances in Neural Information Processing Systems.
  vol.~17. MIT Press (2005)

\bibitem{Kaizhu2009}
Huang, K., Ying, Y., Campbell, C.: Gsml: A unified framework for sparse metric
  learning. In: Proceedings of the 2009 Ninth IEEE International Conference on
  Data Mining (2009)

\bibitem{Duda2001}
O.Duda, R., E.Hart, P., Stork, D.G.: Pattern Classification. A Wiley
  Interscience Publication (2001)

\bibitem{Rakotomamonjy2003}
Rakotomamonjy, A.: Variable selection using svm-based criteria. Journal of
  Machine Learning Research  3,  1357--1370 (2003)

\bibitem{Schultz2004}
Schultz, M., Joachims, T.: Learning a distance metric from relative
  comparisons. In: NIPS (2004)

\bibitem{Shwartz2004}
Shalev-Shwartz, S., Singer, Y., Y.Ng, A.: Online and batch learning of pseudo
  metrics. In: ICML (2004)

\bibitem{weinberger2009distance}
Weinberger, K., Saul, L.: {Distance metric learning for large margin nearest
  neighbor classification}. The Journal of Machine Learning Research  10,
  207--244 (2009)

\bibitem{xing2003dml}
Xing, E., Ng, A., Jordan, M., Russell, S.: {Distance metric learning with
  application to clustering with side-information}. In: Advances in neural
  information processing systems. MIT Press (2003)

\bibitem{Zhu03}
Zhu, J., Rosset, S., Hastie, T., Tibshirani, R.: 1-norm support vector machine.
  In: Neural Information Processing Systems. p.~16. MIT Press (2003)

\end{thebibliography}
\bibliographystyle{splncs03}

 \section*{Appendix}
\subsection*{Showing the equivalence of problem~(\ref{SVM_gamma_sqd}) and problem~(\ref{eq:SVMMetric_org}) (Section \ref{sec:SVM-metric-view}):} 
With the 'symmetry' preference as described after problem~(\ref{SVM_gamma_sqd}), the second constraint of problem~(\ref{SVM_gamma_sqd}) can be replaced by 
the constraint: $ y_i(\vect w^T \vect x_i + b) \geq \gamma /2 $. Moreover,� we can always replace the squared values in problem~(\ref{SVM_gamma_sqd})
by their respective absolute values and get:
\begin{eqnarray}
\label{eq1}
 \max_{\vect w}  \,\, \,\, \gamma &&
s.t.  \,\, \,\,  | \vect w^T (\vect x_i - \vect x_j ) |� \geq \gamma , y_i \neq y_j \nonumber \\
 &&  \,\, \,\, \,\, y_i(\vect w^T \vect x_i + b)\geq \gamma /2, \forall i, \nonumber
\end{eqnarray}
If the constraint � $ y_i(\vect w^T \vect x_i + b) \geq \gamma /2 $ is satisfied then $\vect w^T \vect x_i + b $ and $ y_i $ have the same sign; therefore any two instances $\forall \vect x_i, \vect x_j$ which have different labels, $(y_i \neq y_j)$, will lie on the opposite sides of the hyperplane. Hence:  
\begin{eqnarray}
\label{margin.equip}
 | \vect w^T (\vect x_i - \vect x_j ) |�  &= &| ( \vect w^T \vect x_i + b ) - (\vect w^T \vect x_j + b ) |  \\
 &=&| \vect w^T \vect x_i + b | + | \vect w^T \vect x_j + b |� � \nonumber\\
&=&�  y_i (\vect w^T \vect x_i + b)�  +�  y_j (\vect w^T \vect x_j + b)� \nonumber\\
&\geq& \gamma/2 + \gamma/2 = \gamma�   � \nonumber
\end{eqnarray}
 Therefore the first constraint of problem~(\ref{SVM_gamma_sqd})�  is always satisfied if the constraint�  $ y_i(\vect w^T \vect x_i + b) \geq \gamma /2 $ 
is satisfied, thus~(\ref{SVM_gamma_sqd}) is equivalent to~(\ref{eq:SVMMetric_org}).

\subsection*{Equivalence of (\ref{eq:SVMMetric_org2}) to standard SVM formulation}
We will show  that problem~(\ref{eq:SVMMetric_org2}) is equivalent to standard \svm. In fact~(\ref{eq:SVM-std-margin}) also scales with uniform scaling of $\vect{w}, b$ and to  avoid this problem, $\|\vect{w}\|\gamma$ is fixed to $1$ which lead to the second formula~(\ref{eq:SVM-std}). We will show that the two ways of avoiding scaling problem, by forcing $\|\vect{w}\|\gamma = 1$ or by forcing $\|\vect w\|=1$ are equivalent. 
Indeed, another way to avoid the scaling problem is to find a quantity which is invariant to the scaling of $\vect{w}$.
Let $\gammanewsvm = t \|\vect{w}\|_p$, hence $t: 0\mapsto \infty$, and lets fix $\|\vect{w}\|_p = 1$. Then let 
$P$ be the feasible set of $\vect w$ which satisfies $(\|\vect{w}\|_p = 1 $ \textit{and} $y_i(\vect{w}^T \vect{x}_i + b) \geq 0, \forall i)$, and $Q=\{\vect w|\|\vect{w}\|_p = 1 \textit{ and } 
y_i(\vect{w}^T \vect{x}_i + b) \geq t \|\vect{w}\|_p),  \forall i\}$  Notice that, if $t=0$ then $Q\equiv P$, if $t>0$ then 
$Q \subseteq P$, and if $t> t_{max}$ then $Q$ will be empty. 
For another value of $\|\vect{w}\|_p$, $\|\vect{w}\|_p = \lambda$, the corresponding feasible sets are 
$P_{\lambda}$ and $Q_{\lambda}$. There is a one to one mapping from $P$ to $P_{\lambda}$, and from $Q$ to $Q_{\lambda}$, 
and the $t_{{max}_{\lambda}}$ which makes $Q_{\lambda}$ empty is the same as $t_{max}$. So $t_{max}$ is 
invariant to the scaling of $\vect{w}$. Therefore (\ref{eq:SVMMetric_org2}) 
is  equivalent to:
\begin{eqnarray}
\label{eq:t-invariant}
\max_{\vect{w},b, t} && t \\
s.t. && y_i(\vect{w}^T \vect{x}_i + b) \geq t\|\vect{w}\|_p, \forall i ,\ t\|\vect{w}\|_p = 1 \nonumber
\end{eqnarray}
The value of the geometric margin here is fixed to $1/\sqrt{d}$ while in standard \svm\ the geometric margin is $\gammasvm=1/\|\vect{w}\|^2_2$.
Using the $l_2$ norm of $\vect{w}$, we get a formulation which is equivalent to that of the hard margin \svm\ given in (\ref{eq:SVM-std}). Using the $l_1$ norm of $\vect{w}$, we get the 1-norm \svm\ \cite{Zhu03}.

\end{document}